\pdfoutput=1
\documentclass[11pt]{article}
\usepackage[preprint]{acl}
\usepackage{times}
\usepackage{latexsym}
\usepackage[T1]{fontenc}
\usepackage[utf8]{inputenc}
\usepackage{microtype}
\usepackage{inconsolata}
\usepackage{graphicx}

\usepackage{colortbl}
\usepackage{color,soul}
\usepackage{booktabs}
\usepackage{dcolumn}
\usepackage{subfig}
\usepackage{hyperref}
\usepackage{multirow}
\usepackage{listings}
\usepackage{array}
\definecolor{lightgray}{gray}{0.95}
\usepackage{xcolor}
\definecolor{red}{HTML}{CD5C5C}
\usepackage{tcolorbox}

\title{Lost in Space: Finding the Right Tokens for Structured Output}

\author{Sil Hamilton \and David Mimno \\
  Department of Information Science \\
  Cornell University \\
  \texttt{\{srh255,mimno\}@cornell.edu}}

\begin{document}
\maketitle
\begin{abstract}
General-purpose language models are trained to produce varied natural language outputs, but for some tasks, like annotation or classification, we need more specific output formats.
LLM systems increasingly support structured output, which enforces formats by sampling tokens according to a grammar --- but also unpredictably reduces downstream performance.
Are there systematic differences between grammars that appear semantically (and often visually) similar to humans?
To answer this, we test four popular model families with five varying output formats on four common NLP benchmarks. 
We find all models perform most accurately when guided to use formats respecting convention, such as letters for multiple choice and real numbers for numerical prediction.
Performance also improves by 5\%-10\% when guiding models to return tokens incorporating leading whitespace, with smaller models benefiting the most.
We find leading whitespace helps models avoid structural deficiencies in subword token representations.
We finally present best practices for researchers using language models as zero-shot classifiers with structured output.
\end{abstract}

\section{Introduction}
\begin{figure}[t]
\begin{tcolorbox}[
    width=\linewidth,
    height=2.6in,
    colframe=gray,
    colback=white,
    boxrule=0.5pt,
    arc=2pt,
]
{\sffamily
\small

\vspace{0.5em}
{\textcolor{black}{\textbf{Prompt:}}

\vspace{0.3em}

\emph{Rate the similarity of these words from 1 to 5: ``train'' and ``underground''. Use only the following characters:}}

\vspace{1.3em}

\textbf{Tokens decoded as digits:}

\vspace{0.3em}

\fbox{\texttt{1}} \hspace{0.2em}
\fbox{\texttt{2}} \hspace{0.2em}
\fbox{\texttt{3}} \hspace{0.2em}
\fbox{\texttt{4}} \hspace{0.2em}
\fbox{\texttt{5}}
\hspace{5.75em} \textcolor{black}{→}
\hfill 3

\vspace{1.3em}

\textbf{Tokens decoded as digits with whitespace:}

\vspace{0.3em}

\fbox{\texttt{\textvisiblespace1}} \hspace{0.2em}
\fbox{\texttt{\textvisiblespace2}} \hspace{0.2em}
\fbox{\texttt{\textvisiblespace3}} \hspace{0.2em}
\fbox{\texttt{\textvisiblespace4}} \hspace{0.2em}
\fbox{\texttt{\textvisiblespace5}}
\hfill \textcolor{black}{→}
\hfill 5

\vspace{1.1em}
\begin{tcolorbox}[
    colback=lightgray,
    coltext=black,
    colframe=lightgray,
    boxrule=0.5pt,
    left=1pt,
    arc=2pt,
    right=1pt,
    top=3pt,
    bottom=3pt,
]
{
\begin{center}
Llama 3.1 picks different tokens depending on whether given tokens incorporate whitespace.
\end{center}
}
\end{tcolorbox}
}

\end{tcolorbox}
\caption{Llama 3.1 \texttt{8b} provides different answers to a question from the MEN Test Collection when constrained to produce digit tokens \textit{with} and \textit{without} a leading space. Sampled with temperature set to 0.}
\label{fig:llama}
\end{figure}
Language model outputs are flexible, which makes them effective for a wide range of applications, but this can also be a liability.
While language models are often used to emit free text, e.g. writing a story, other important applications require their output match specific formats like class labels and domain-specific languages \citep{wangGrammarPromptingDomainSpecific, Geng_Josifoski_Peyrard_West_2023}.
Rigid formats make it easier to parse responses, helping users avoid wasting queries on potentially invalid stochastic output.

Multiple choice options are the simplest example, but users are increasingly using more flexible (and more failure-prone) output format restrictions.
Reinforcement learning from human feedback (RLHF) and instruction tuning can guarantee well-formed output \citep{Ouyang_Wu_Jiang_Almeida_Wainwright_Mishkin_Zhang_Agarwal_Slama_Ray_et_al._2022}, but these methods are time consuming and expensive to implement, and are impractical for the variable and rapid-turnaround scenarios typical of LLM usage.

An increasingly popular method for ensuring output consistency without fine-tuning is selectively sampling tokens valid according to a formal language \citep{Shin_Lin_Thomson_Chen_2021, Wang_Wang_Wang_Cao_Saurous_Kim_2023}.
Structured output, powered by grammar-constrained decoding (GCD), is regularly implemented in server software for running LLM inference \citep{Gerganov_2024, Microsoft_2024, Rickard_2024}.
Some servers allow users to specify grammars with regular languages \citep{Willard_Louf_2023, Microsoft_2024}, while others allow users to write context-free grammars in Backus–Naur form  \citep{Chomsky_1956, Knuth_1964, Gerganov_2024}.

But despite this popularity, enabling GCD appears to degrade downstream task accuracy \citep{Willard_Louf_2023, Park_Wang_Berg_2024, Beurer-Kellner_Fischer_Vechev_2024, Geng_Josifoski_Peyrard_West_2023, Le_Chen_Ritter_Xu_2024}.
We hypothesize that the reduced performance of GCD is not due to GCD itself, but to non-optimal grammars.
For example, \autoref{fig:llama} shows Llama 3.1 produces different results when constrained by two different output token sets, one with leading spaces and another without.

We investigate this phenomenon by comparing the performance achieved when using one of five distinct token sets across four different data labeling scenarios with four popular families of pre-trained language models.
We present three key contributions. 
First, although all models vary considerably between formats for a given task, they often agree on a preferred format, such as real numbers for prediction tasks.
Second, important distinctions may not be obvious to humans.
We find that tokens incorporating leading whitespace consistently improve performance, echoing results for syntactic tokens \citep{Pimentel_Meister_2024, Oh_Schuler_2024}.
We offer initial explanation for this difference using evidence in the embedding weights of our models of interest.
We finally propose a set of strategies for selecting target tokens when specifying grammars for GCD libraries. 
These strategies can help achieve better performance when using structured output, particularly for users of smaller, less computationally intensive models.

\section{Related Work}

\paragraph{Grammar-constrained decoding failure modes.}
Grammar-model misalignment is a recognized issue with grammar-constrained decoding algorithms.
Existing solutions generally involve modifying GCD algorithms to avoid misalignment by adding additional constraints.
\citet{Park_Wang_Berg_2024} propose an adaptive algorithm where beam search discovers valid \emph{and} optimal token sequences according to the grammar.
They report some success in generating valid sequences, but note accuracy is not improved in all respects.
\citet{Beurer-Kellner_Fischer_Vechev_2024} propose ``DOMINO,'' a ``minimally invasive'' algorithm, pre-computing allowed sequences of tokens before dynamically selecting certain paths according to probability mass.
While these approaches show promise, both propose non-trivial adjustments to how GCD is commonly implemented in software.
This limits their downstream applicability, and does not provide a generalizable token alignment strategy for common inference servers \emph{as they are presently implemented}.

\paragraph{Non-exchangeable token embeddings.}
Artifacts in LLM training can result in tokens that appear semantically similar to humans but whose latent representations differ to a great extent.
\citet{Wen-Yi_Mimno_2023} explore trained token representations in the embedding weights of highly multilingual LLMs, observing significant variation in how models represent semantically similar tokens in the embedding space.
\citet{Zhang_Lu_Tran_Schuster_Metzler_Lin_2024} report similar behavior in monolingual LLMs, noting asymmetries in token categories such as pronouns and nouns.
Neither study assesses the downstream impact of this behavior.

\paragraph{Leading whitespace tokens.}
Tokenizers descending from SentencePiece duplicate certain tokens: once with leading whitespace to capture instances where the token begins a word, and again without whitespace \citep{Kudo_Richardson_2018}.
For example, the Llama 3 tokenizer can represent ``culture'' either as the token \texttt{culture} (index key $70905$) or \texttt{␣culture} ($7829$).
These tokens are distinct despite decoding to strings nearly identical to humans.

\citet{Pimentel_Meister_2024} note tokenizers making use of LW tokens can complicate computing token surprisal.
\citet{Oh_Schuler_2024} similarly find models trained with LW tokens are prone to an increased likelihood of failing to learn word boundaries, given their tokenizers explicitly penalize the appearance of space tokens.
Neither work estimates the impact tokens may have on downstream tasks, nor whether LW tokens can converge with non-LW tokens in training.

\section{Methods}

If we can dependably elicit grammar preferences, we can avoid extending existing GCD-enabled inference servers as is routinely recommended by the literature \citep{Geng_Josifoski_Peyrard_West_2023, Park_Wang_Berg_2024, Beurer-Kellner_Fischer_Vechev_2024}.
In this section we describe our method for systematically discovering these GCD-optimal token sets.

\renewcommand{\arraystretch}{1.1}
\begin{table*}[t]
\centering
\small
\begin{tabular}{lllll}
\textbf{Category} & \textbf{Variant} & \textbf{Token Set} & \textbf{Maximum Length} & \textbf{Example} \\
\midrule
\multirow{2}{*}{Integer} & Integer, \emph{numeric} & $\{1,2,...,10\}$ & 1 token & 7 \\
 & Integer, \emph{word} & \{One, Two, ..., Ten\} & 1 token & ``Seven'' \\
\cmidrule(lr){1-5}
\multirow{2}{*}{Real Number} & Real, \emph{numeric} & $\{0.1,...,1\}$ & $\leq3$ token(s) & 0.1 \\
 & Real, \emph{word} & \{Zero point zero, ..., One\} & $\leq4$ token(s) & ``Zero point one'' \\
\cmidrule(lr){1-5}
\multirow{2}{*}{Percentage} & Percentage, \emph{numeric} & $\{0\%$, ..., $100\%\}$ & 2 tokens & 10\% \\
 & Percentage, \emph{word} & \{Zero, ..., One hundred; percent\} & $\leq4$ token(s) & ``Ten percent'' \\
\cmidrule(lr){1-5}
\multirow{2}{*}{Boolean} & Binary, \emph{numeric} & $\{1, 0\}$ & 1 token & 1 \\
 & Binary, \emph{word} & \{True, False\} & 1 token & ``True'' \\
\cmidrule(lr){1-5}
\multirow{2}{*}{Likert} & Likert task, \emph{numeric} & $\{1,2,3,4,5\}$ & 1 token & 4 \\
 & Likert task, \emph{word} & \{Strongly disagree, ..., Strongly agree\} & $\leq3$ token(s) & ``Somewhat agree'' \\
\end{tabular}
\caption{Target token sets considered in our study, noting the value range and specifying the maximum length of a valid sequence with Llama 3.1 followed by an example.}
\label{tab:target-labels}
\end{table*}

\paragraph{Evaluation.} 
\label{par:evaluation}
We assess the viability of certain token formats against four widely-used NLP benchmarks that involve quantitative outputs.\footnote{Common multiple choice benchmarks (e.g. MMLU) are subject to token influence (see \autoref{tab:benchmark-analysis}), but do not require GCD as responses can be read straight from model logits. Compare with prediction tasks where responses may be composed of two to three tokens, e.g. 12\% being composed of the tokens \textit{[1], [2], [\%]}.}
Our first two benchmarks are the Semantic Textual Similarity Benchmark \citep{cerSemEval2017Task12017} and the MEN Test Collection \citep{Bruni_Tran_Baroni_2014}. 
Both are prediction tasks asking annotators to rate the semantic similarity of two sentences on a scale between two numbers.
Our next two benchmarks are \emph{binary detection} tasks asking annotators to assess whether a given phenomenon is present: Quora Question Pairs \citep{DataCanary_hilfialkaff_Jiang_Risdal_Dandekar_tomtung_2017} and ToxicChat \citep{Lin_Wang_Tong_Wang_Guo_Wang_Shang_2023}. 

We randomly sample 1,000 problems from each benchmarks for a total of 4,000 problems.
We then specify a standard task prompt template valid for problems sampled from all four datasets.
We present example task prompts for all four benchmarks in \autoref{sec:appendix}.

\paragraph{Baseline measures.}
We contextualize our LLM results using two baseline measures.
Given two of our four benchmarks target semantic similarity as a real number, we make use of \verb.BERTScore. to provide a baseline BERT-based semantic similarity score for all problems \citep{Zhang_Kishore_Wu_Weinberger_Artzi_2020}.
This system was trained to perform semantic similarity scoring, conveniently serving as an upper bound to compare our zero-shot prompting strategies against.
We estimate the lower bound by measuring the Levenshtein distance \citep{Levenshtein_1966} between both choices in all problems, with the expectation these scores will trend lower given all benchmarks appear only lightly correlated with string edit distance.

We intend neither measure to be state-of-the-art.
Rather, BERTScore represents the performance of a reasonable language model, while the edit distance of each problem is a simple method expected to outperform random guessing by some margin.

\paragraph{Model selection.}
To assess whether grammar-dependent degradations are consistent between LLMs, we evaluate eight models of varying size.\footnote{We do not include proprietary models in our assessment because existing APIs, e.g. that of OpenAI, do not enable defining context-free grammars.}
We select these eight from four popular families of instruction-tuned LLMs: Llama 3.1 \verb.8b./\verb.70b., Qwen 2.5 \verb.7b./\verb.72b., Gemma 2 \verb.9b./\verb.24b. and Phi 3 \verb.3b./\verb.14b. \citep{Dubey_Jauhri_Pandey_et_al._2024, qwenQwen25TechnicalReport2025, teamGemma2Improving2024, abdinPhi3TechnicalReport2024}.
All models were trained in the last two years and feature pre-training sets ranging between 3.3 and 18 trillion tokens.
We select these model families for three reasons: their high performance relative to proprietary models, their high compatibility with GCD-enabled inference servers, and their popularity with researchers for certain downstream tasks such as classification and annotation, e.g. \citet{Bamman_Chang_Lucy_Zhou_2024}.

\paragraph{Model inference.}
All models are run with model-default temperature and \verb.Top_p. parameters.
We set context length to 512 tokens in all cases.
We obtain models from their official repositories, quantizing each to \verb.q8_0. with scripts provided by \citet{Gerganov_2024} to maximize local inference speed.\footnote{Prior research suggests reducing weight precision to an average of no fewer than eight bits ensures quantized performance remains comparable to unquantized models \citep{Dettmers_Lewis_Belkada_Zettlemoyer_2022}.}
We host the models with \verb,llama.cpp, (compiled with CUDA support) and run the server on a system equipped with two Nvidia RTX A6000 GPUs.
We use the GBNF GCD implementation included with \verb,llama.cpp, without modification.\footnote{Examples of GBNF grammars can be found \href{https://github.com/ggerganov/llama.cpp/blob/master/grammars/README.md}{here}.}

\paragraph{Token output formats.}
\label{par:Target Label Ranges}
\renewcommand{\arraystretch}{1.1}
\begin{table}[t]
\centering
\small
\begin{tabular}{l*{4}{c}}
 & \multicolumn{2}{c}{\textbf{BERTScore}} & \multicolumn{2}{c}{\textbf{Levenshtein}} \\
\cmidrule(lr){2-3} \cmidrule(lr){4-5}
Benchmark & Corr. & MSE & Corr. & MSE \\
\midrule
MEN & 0.768 & 397 & $-0.055$ & 3,736 \\
QUORA & 0.609 & 2,574 & $-0.323$ & 8,106 \\
STSB & 0.824 & 379 & $-0.107$ & 8,076 \\
TOXICCHAT & $-0.081$ & 4,248 & $-0.014$ & 267,889 \\
\midrule
$\mathbf{\bar{x}}$ & 0.530 & — & $-0.125$ & — \\
\end{tabular}
\caption{Baseline correlations and MSE for each STS benchmark considered in this paper. Note semantic similarity consistently outperforms Levenshtein distance.}
\label{tab:baseline-correlations}
\end{table}
Current GCD implementations have users specify output formats with CFGs meaning many possible output formats exist.

We consider five commonly used output formats, shown in \autoref{tab:target-labels}.
These categories include integers between 1 and 5 inclusive, real numbers in the closed interval $[0,1]$, percentages, binaries, and 5-point Likert scales.
We further consider three additional format variables:

\begin{itemize}
    \item \emph{With/without leading newline}. Chat templates for RLHF-tuned models can implicitly expect a new line following the beginning of the assistant turn.
    \item \emph{With/without leading space}. Previous research has identified tokens incorporating a leading space character as being potentially preferred by models.
    \item \emph{As digits/words}. All token categories can be represented in either numeric and English word tokens.
\end{itemize}

We vary across datasets, token families, model families, model sizes, treatments, and our baseline measures.
This results in 1,280,000 classifications.

\section{Results}
\label{sec:results}
\renewcommand{\arraystretch}{1.1}
\begin{table}[t]
\centering
\small
\begin{tabular}{l*{5}{D{.}{.}{1.3}}}
\multicolumn{6}{c}{\textbf{Correlations --- Small Models}} \\
\midrule
Model & \multicolumn{1}{c}{Integer} & \multicolumn{1}{c}{Real} & \multicolumn{1}{c}{Percent} & \multicolumn{1}{c}{Binary} & \multicolumn{1}{c}{Likert} \\
\midrule
Llama & -0.160 & \multicolumn{1}{c}{\textbf{0.420}} & 0.046 & 0.199 & 0.191 \\
Gemma & 0.181 & \multicolumn{1}{c}{\textbf{0.537}} & 0.346 & 0.118 & 0.138 \\
Phi & 0.436 & \multicolumn{1}{c}{\textbf{0.505}} & 0.422 & 0.402 & 0.453 \\
Qwen & 0.331 & \multicolumn{1}{c}{\textbf{0.568}} & 0.372 & 0.344 & 0.466 \\
\midrule
$\mathbf{\bar{x}}$ & 0.197 & \multicolumn{1}{c}{\textbf{0.507}} & 0.296 & 0.266 & 0.312 \\
\addlinespace[1em]
\multicolumn{6}{c}{\textbf{Correlations --- Large Models}} \\
\midrule
Model & \multicolumn{1}{c}{Integer} & \multicolumn{1}{c}{Real} & \multicolumn{1}{c}{Percent} & \multicolumn{1}{c}{Binary} & \multicolumn{1}{c}{Likert} \\
\midrule
Llama & 0.151 & \multicolumn{1}{c}{\textbf{0.492}} & 0.077 & 0.306 & 0.311 \\
Gemma & 0.155 & \multicolumn{1}{c}{\textbf{0.476}} & 0.110 & 0.095 & -0.006 \\
Phi & 0.362 & 0.412 & 0.376 & 0.335 & \multicolumn{1}{c}{\textbf{0.413}} \\
Qwen & 0.429 & 0.609 & \multicolumn{1}{c}{\textbf{0.659}} & 0.472 & 0.581 \\
\midrule
$\mathbf{\bar{x}}$ & 0.274 & \multicolumn{1}{c}{\textbf{0.497}} & 0.306 & 0.302 & 0.325 \\
\end{tabular}
\caption{The \textit{real number} format has the highest Spearman correlation with human labels. Larger models show less differentiation between formats. 
}
\label{tab:model-correlations}
\end{table}
We first evaluate the effectiveness of different output formats by measuring the correlation between the numeric value of a model's output prediction and the human label.
Because the model prediction formats and human labels do not necessarily map to real numbers (e.g. Likert), we report Spearman rank correlation. Pearson correlations, however, are typically within 0.01 from the Spearman values.

\paragraph{Baseline correlations.}
We first evaluate the two baseline measures (\verb.BERTScore. and Levenshtein distance) over all four benchmarks.
We present correlations and MSE in \autoref{tab:baseline-correlations}.
As expected, \verb.BERTScore., a system specifically trained for semantic similarity, shows strong correlation on our two semantic similarity benchmarks ($r\!=\!0.755$ and $r\!=\!0.839$). 
It performs less well on Quora Pairs  $r\!=\!0.540$ and near random on ToxicChat $r\!=\!-0.061$.
These values serve as an indication of the capabilities and limitations of systems fine-tuned for particular tasks.
On the other extreme, the Levenshtein distance baseline performs poorly in all cases, with mean correlation of $r\!=\!-0.094$.
This result replicates \citet{Lin_Wang_Tong_Wang_Guo_Wang_Shang_2023}, whose task involves classifying the presence of toxic language and is thus explicitly not measuring semantic similarity.
This result indicates that the tasks are indeed hard and cannot be solved by na{\"i}ve algorithms. 

\paragraph{Real numbers perform best for prediction.}
\autoref{tab:model-correlations} shows correlations between model output and human labels broken down by model and token set.\footnote{We do not provide the corresponding MSE for our non-baseline correlations due to the inherent limited range of certain token sets, e.g. binary scores and Likert numbers, being a confounding factor.}
Model scores were obtained without any additional formatting, such as newline after the beginning of the assistant turn or leading space before the classification token.
Results are the mean of observations across all benchmarks.\footnote{We emphasize here that absolute correlation values are less meaningful than the relative scores across model types.}
Real numbers perform best across all models, with a minimum Spearman rank correlation coefficient of $\rho\!=\!0.420$ and a maximum of $\rho\!=\!0.609$.
We conversely find integers and binaries perform worst, e.g. $\rho\!=\!-0.160$ for integers on Llama 3.1 \verb.8b..
Likert proves the least reliable, with the label range yielding a correlation of $\rho\!=\!-0.006$ with Gemma 2 \verb.27b. and $\rho\!=\!0.581$ using Qwen 2.5 \verb.72b..

These results indicate that there are consistent, substantial differences between output formats, and that real-valued outputs are consistently the best format across all models in prediction tasks. 
We offer further exploration in \autoref{par:tokens}.

\paragraph{Format differences are strongest for smaller models.}
Averaging over all formats, large models have higher correlation than small models, at $0.341\pm0.28$ and $0.316\pm0.39$ respectively.
This trend is due mostly, however, to the Llama and Qwen families.
Phi 3 and Gemma 2 show little improvement or even reduction in performance. For Likert scores, Gemma goes \textit{down} from  $\rho\!=\!0.138$ to $\rho\!=\!-0.006$ between small and large models.
We investigate this phenomenon more closely in \autoref{par:tokens}.
The difference between real values and the other formats is also less pronounced for large models than for small models.
The global mean values likewise disguise variations in per-token set performance.
While increasing parameter count improves correlation for most formats, we find integers exhibit a mean increase of less than 3.5\%.
This intra-set variation is most pronounced in Gemma 2 \verb.27b., which yields lower mean correlation scores for the percent, binary, and Likert formats.

\paragraph{Whitespace matters.}
\renewcommand{\arraystretch}{1.1}
\begin{table}[t]
\centering
\small
\begin{tabular}{l*{4}{c}}
\multicolumn{5}{c}{\textbf{Treatment --- Correlation Deltas ($\Delta\rho$)}} \\
\midrule
Condition & \multicolumn{1}{c}{Llama} & \multicolumn{1}{c}{Gemma} & \multicolumn{1}{c}{Phi} & \multicolumn{1}{c}{Qwen} \\
\midrule
with\_newline & \textbf{0.056} & \textbf{0.138} & 0.006 & 0.037 \\
with\_space & 0.037 & 0.016 & \textbf{0.036} & 0.011 \\
as\_word & -0.497 & -0.510 & -0.451 & -0.815 \\
as\_large & 0.005 & -0.019 & 0.014 & \textbf{0.059} \\
\midrule
$\mathbf{\bar{x}}$ & -0.100 & -0.094 & -0.099 & -0.177 \\
\end{tabular}
\caption{Treatment effect deltas. Positive values mean treatment improves correlation with ground truth.}
\label{tab:treatment-correlations}
\end{table}

\renewcommand{\arraystretch}{1.1}
\begin{table*}[!htb]
\centering
\small
\begin{tabular}{l@{\hspace{20pt}}r@{\hspace{16pt}}l@{\hspace{16pt}}l@{\hspace{16pt}}l@{\hspace{16pt}}l@{\hspace{16pt}}l}
\textbf{Benchmark} & \textbf{Stock} & \textbf{With Newline} & \textbf{With Space} & \textbf{As Integer} & \textbf{As Real} & \textbf{As Word} \\
\midrule
ARC-C & 59 & \textbf{77} \, {\color{green!50!black}(+30\%)} & 63 \, {\color{green!50!black}(+7\%)} & 32 \, {\color{red!70!black}(-46\%)} & 22 \, {\color{red!70!black}(-63\%)} & 54 \, {\color{red!70!black}(-8\%)} \\
BoolQ & 38 & \textbf{58} \, {\color{green!50!black}(+53\%)} & 41 \, {\color{green!50!black}(+8\%)} & 56 \, {\color{green!50!black}(+47\%)} & 46 \, {\color{green!50!black}(+21\%)} & 32 \, {\color{red!70!black}(-16\%)} \\
CommonsenseQA & 39 & \textbf{75} \, {\color{green!50!black}(+92\%)} & 62 \, {\color{green!50!black}(+59\%)} & 29 \, {\color{red!70!black}(-26\%)} & 20 \, {\color{red!70!black}(-49\%)} & 59 \, {\color{green!50!black}(+51\%)} \\
HellaSwag & 43 & \textbf{62} \, {\color{green!50!black}(+44\%)} & 37 \, {\color{red!70!black}(-14\%)} & 28 \, {\color{red!70!black}(-35\%)} & 24 \, {\color{red!70!black}(-44\%)} & 43 \, {\color{gray}(+0\%)} \\
MMLU & 50 & \textbf{62} \, {\color{green!50!black}(+24\%)} & 52 \, {\color{green!50!black}(+4\%)} & 34 \, {\color{red!70!black}(-32\%)} & 24 \, {\color{red!70!black}(-52\%)} & 48 \, {\color{red!70!black}(-4\%)} \\
OpenBookQA & 58 & \textbf{79} \, {\color{green!50!black}(+36\%)} & 66 \, {\color{green!50!black}(+14\%)} & 36 \, {\color{red!70!black}(-38\%)} & 23 \, {\color{red!70!black}(-60\%)} & 66 \, {\color{green!50!black}(+14\%)} \\
Winogrande & 53 & \textbf{58} \, {\color{green!50!black}(+9\%)} & 56 \, {\color{green!50!black}(+6\%)} & 49 \, {\color{red!70!black}(-8\%)} & 51 \, {\color{red!70!black}(-4\%)} & 54 \, {\color{green!50!black}(+2\%)} \\
\midrule
$\mathbf{\bar{x}}$ & 49 & \textbf{67} \, {\color{green!50!black}(+37\%)} & 54 \, {\color{green!50!black}(+10\%)} & 38 \, {\color{red!70!black}(-22\%)} & 30 \, {\color{red!70!black}(-39\%)} & 51 \, {\color{green!50!black}(+4\%)} \\
\end{tabular}
\caption{Llama 3.1 8b's performance on seven common reasoning benchmarks after format and treatments have been applied. As predicted, leading whitespace positively impacts response accuracy. Bold values indicate best performance per benchmark. Percentages show change relative to stock baseline.}
\label{tab:benchmark-analysis}
\end{table*}
To capture what other factors matter beyond model size and token set, we vary several additional format variables described in \autoref{par:Target Label Ranges}.
We calculate the values in \autoref{tab:treatment-correlations} by taking the delta between each model family's correlation with and without the treatment, averaged across all benchmarks.
A positive delta indicates the treatment improved model performance relative ground truth.
This process reveals a number of key trends.
First, \verb.as_word. dramatically reduces performance.
That is, enforcing numeric values as English words reduces correlation for all models: $\Delta\rho\!=\!-0.497$ for Llama, $\Delta\rho\!=\!-0.510$ for Gemma, $\Delta\rho\!=\!-0.451$ for Phi, and $\Delta\rho\!=\!-0.815$ for Qwen.
Second, although we do find an improvement from increasing model parameter count \verb.as_large., this improvement is less significant than including either \verb.with_newline. (leading whitespace in the token immediately prior), and \verb.with_space. (leading whitespace in the token itself).
Doing so results in an even greater performance increase across the Llama, Gemma, and Qwen families.
We discuss this phenomenon further in \autoref{par:LW}.

\section{Analysis}
We observe substantial and consistent variation between output formats. 
In this section we attempt to explain why these formats are so different.

\begin{figure}[t]
\centering
\includegraphics[width=0.49\linewidth]{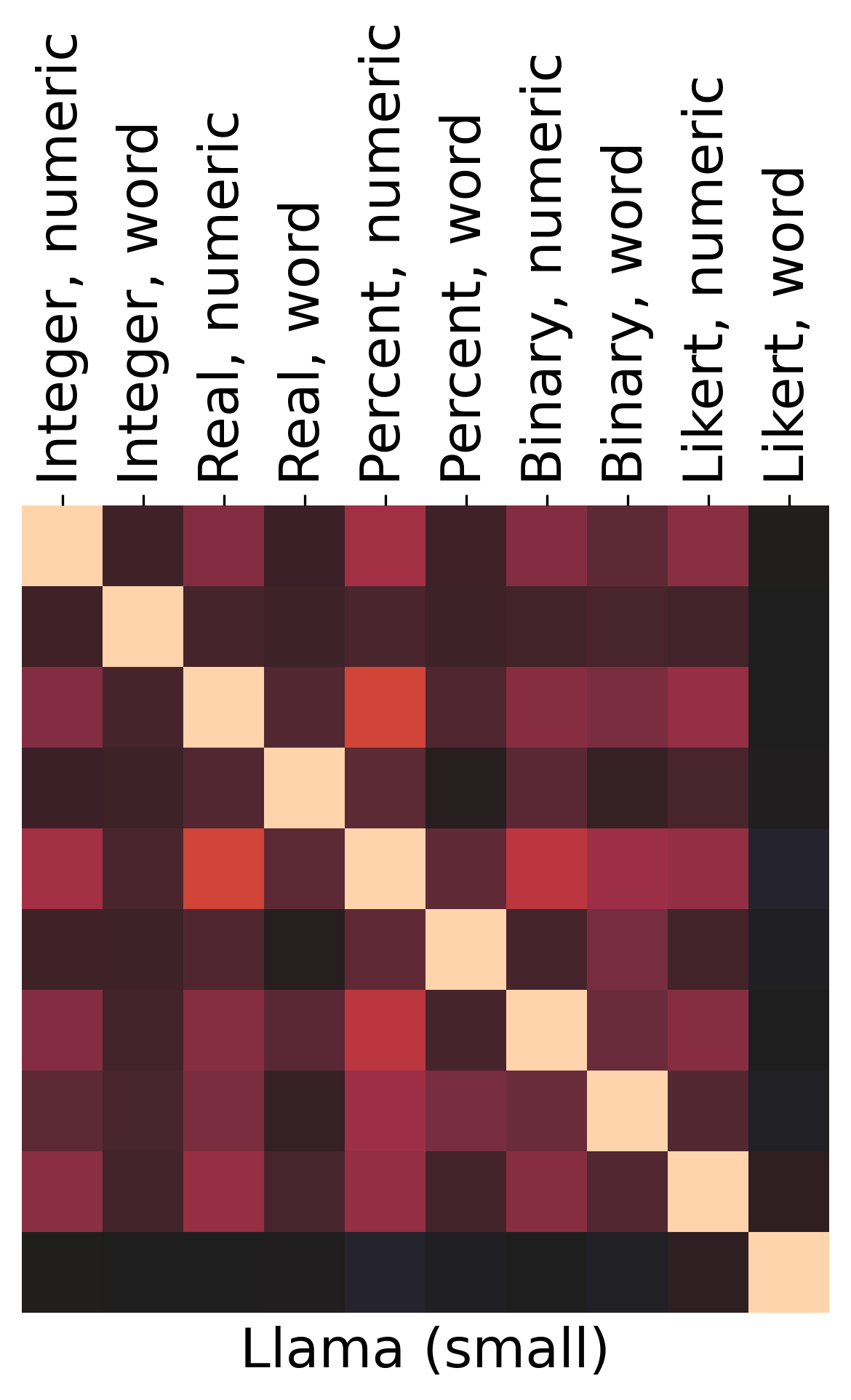}
\includegraphics[width=0.49\linewidth]{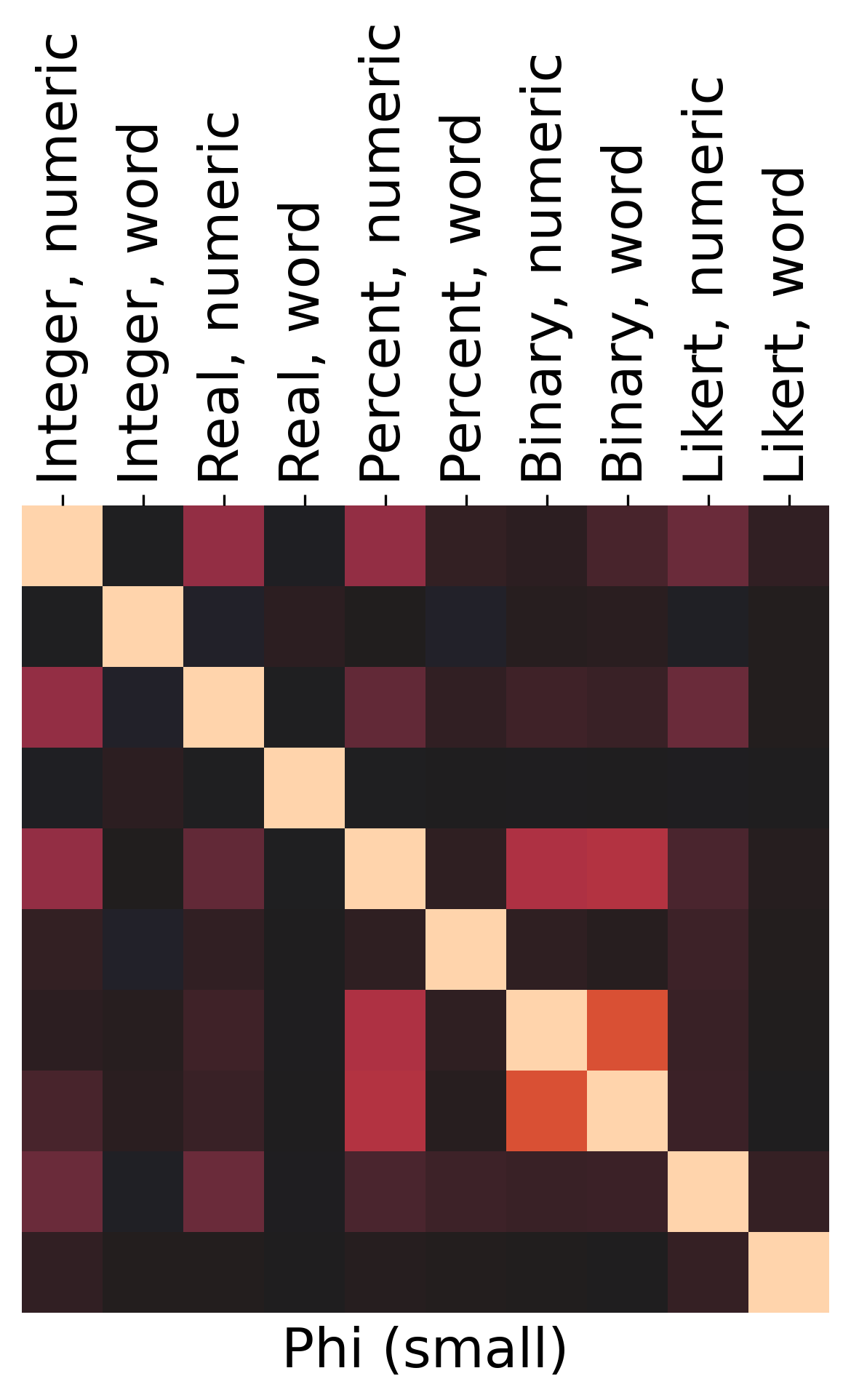}
\includegraphics[width=0.49\linewidth]{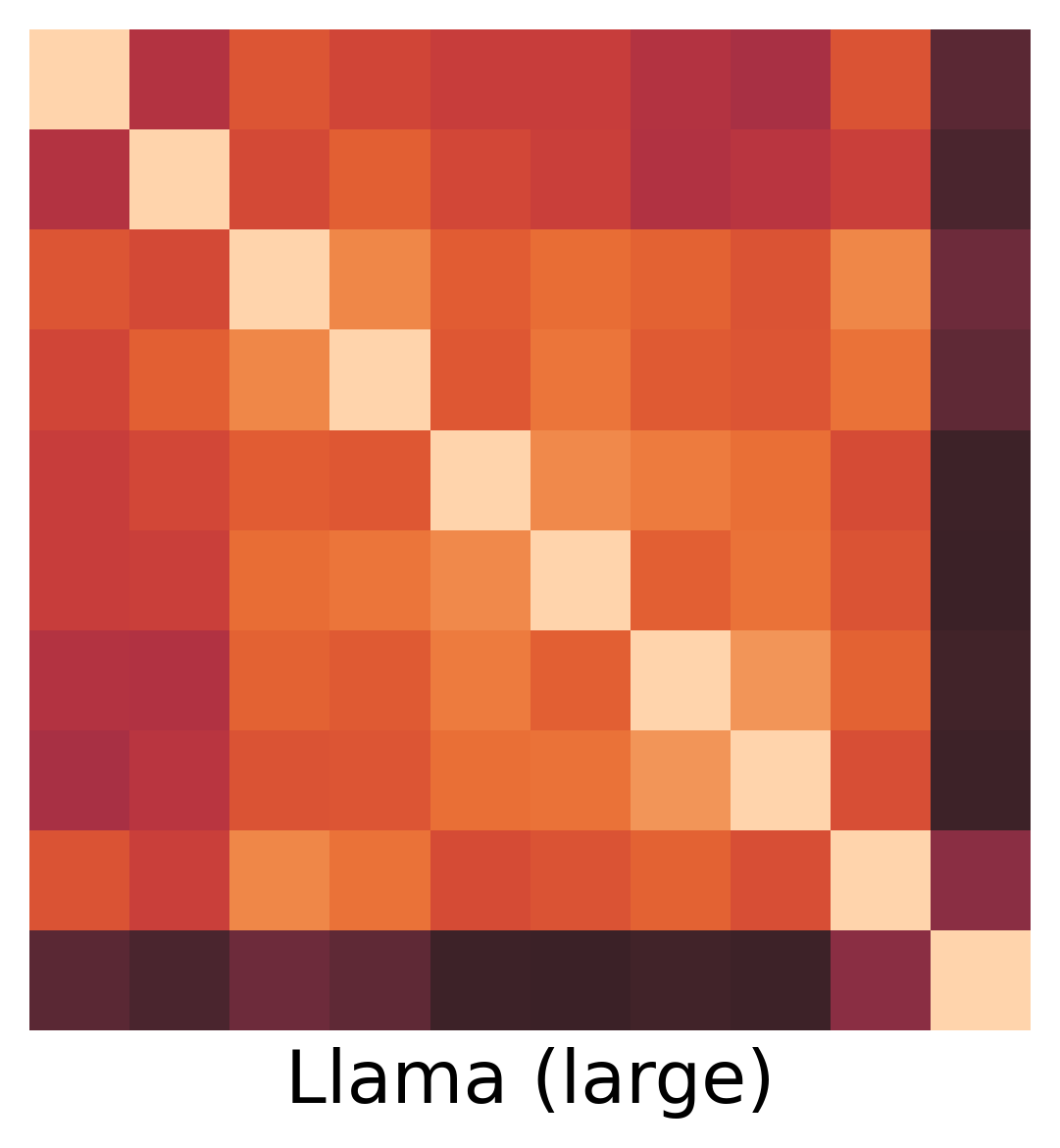}
\includegraphics[width=0.49\linewidth]{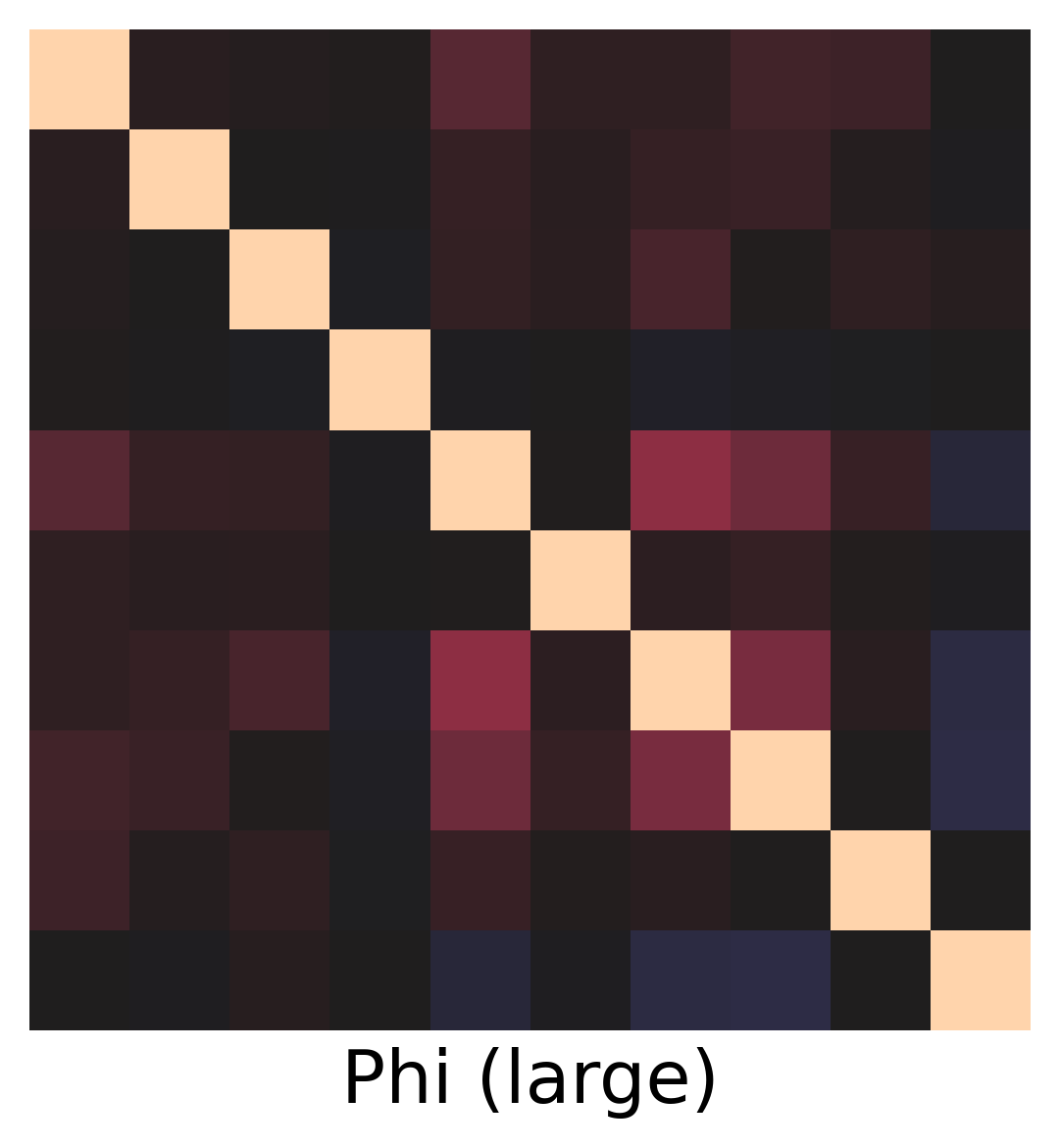}
\caption{Intra-model token set correlation matrices for Llama and Phi family models. Brighter colors indicate higher correlation. Scaling behavior differs between Llama and Phi models.}
\label{fig:heatmap}
\end{figure}

\paragraph{Results differ by format.}
\label{par:tokens}
How stable are our models' responses across format types?
We demonstrate the correlations between all annotations made with each format type for two model families, Llama and Phi, in  \autoref{fig:heatmap}.
Stable responses with each format should yield a solid color in our correlation matrix --- but we instead find all model families (especially their smaller siblings) demonstrate poor intra-format regularity.
Neither when we represent the same scales (e.g. integers) both numerically and in letters (e.g. the number 5 versus ``five'') do we find correlated annotations.
This result indicates no model in our study has achieved stable token representations for the tasks and token sets tested.
Comparing small and large models does reveal increasing parameter count (and thereby training tokens) decreases differentiation between formats for all pairs considered belonging to Llama, Gemma, and Qwen.
One possible explanation for Phi being an outlier is the large portion of synthetic data used in its training data --- synthetic data can amplify certain signals in noise, increasing the possibility of overfitting.

\paragraph{Downstream impact.}
Do our results hold for downstream tasks, such as classification and reasoning?
If our results in \autoref{sec:results} hold, then we would expect Llama to continue exhibiting non-uniform performance when guided to emit different formats.
To investigate, we ran Llama 3.1 8b on seven common LLM reasoning benchmarks with similar formats and treatments as previously described in \autoref{par:Target Label Ranges}.
We test the following formats: \textit{Stock} uses letter choices (A, B...), \textit{As Integer }uses numbers (1, 2...), \textit{As Real} uses decimal numbers (0.00, 0.33...), \textit{As Word} uses choice labels (``Choice 1'', ``Choice 2''...), and finally \textit{With Space} and \textit{With Newline} add leading whitespace and  newlines, respectively.
We present our findings in \autoref{tab:benchmark-analysis}.
While all the benchmarks are multiple choice, they vary by the number of alternative answers for each question: BoolQ and Winogrande are binary \citep{clarkBoolQExploringSurprising2019, sakaguchiWinoGrandeAdversarialWinograd2019}; ARC-C, MMLU, OpenBookQA, and HellaSwag all present four choices \citep{clarkThinkYouHave2018, hendrycksMeasuringMassiveMultitask2021a, mihaylovCanSuitArmor2018, zellersHellaSwagCanMachine2019}; and CommonsenseQA offers five \citep{talmorCommonsenseQAQuestionAnswering2019}.
We vary the number of options accordingly.
We find Llama 3.1 8b continues to demonstrate uneven performance across formats and treatments, with newlines and leading whitespace improving over stock performance by an average 37\% and 10\%.
Whereas we previously found models preferred real numbers when attempting prediction tasks, here we observe models preferring letters over integers (-22\%) and real numbers (-39\%) when selecting multiple choice answers.
This again suggests models prefer certain formats, with the conventional format of the task (e.g. real numbers for prediction) being a predictor.

\paragraph{Leading whitespace prevalence rates.}
\label{par:LW}

\begin{figure*}[t]
\centering
\includegraphics[width=0.49\linewidth]{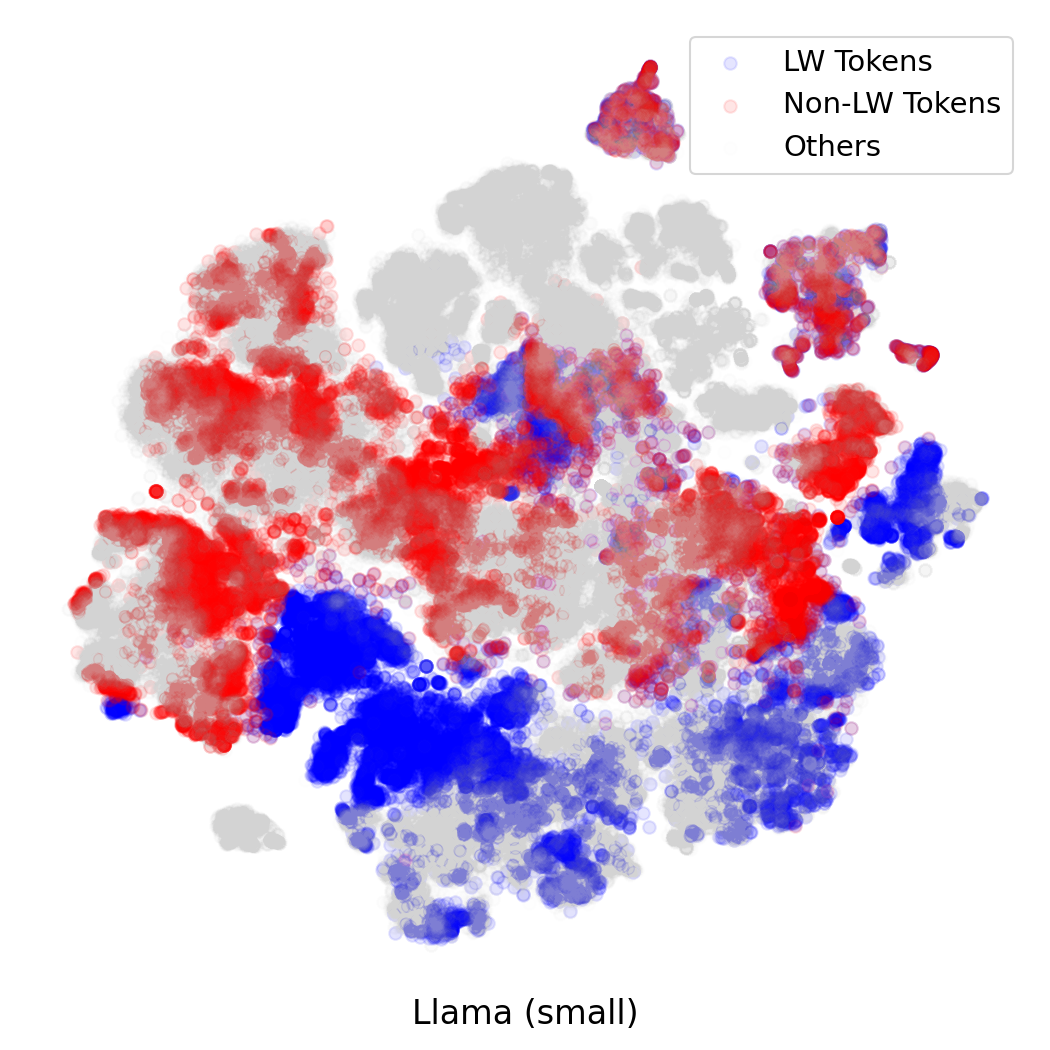}
\includegraphics[width=0.49\linewidth]{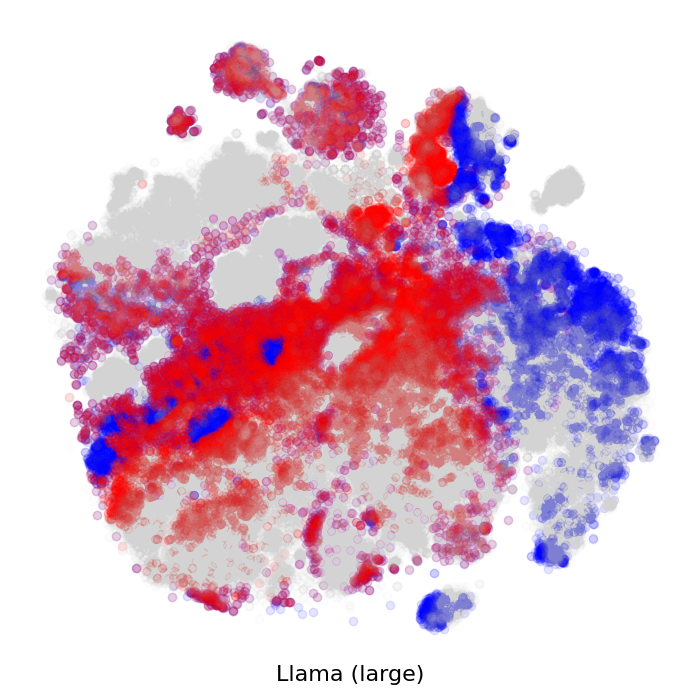}
\caption{LW/non-LW pairs visualized across all models of the Llama family. Larger models tend to use their representation space more efficiently, leading to greater subword token representation similarity.}
\label{fig:tsne}
\end{figure*}

Why do formats including leading whitespace (LW) have better zero-shot performance than otherwise identical formats lacking leading whitespace space?
We investigate the degree LW tokens differ from their non-LW pairs in the embedding space of each of our model families to answer this question.\footnote{We identify LW/non-LW pairs by appending whitespace to all tokens in each vocabulary, tokenizing the resulting string. We then filter for concatenations resulting in a single token.}
We first obtain the tokenizers for each of our model families and extract all LW/non-LW token pairs from their corresponding vocabularies.
We find between 22$\%$ and 32$\%$ of tokens in the Phi, Llama and Qwen vocabularies participate in a pair, while 40$\%$ of the Gemma vocabulary participates in a pair.
Put another way, the lower bound observed has one in ten tokens having a twin — and the upper bound is every one in 2.5.

\paragraph{Leading whitespace shared semantics.}
Do these pairs share semantic representations?
We take the token index keys for each pair and obtain the corresponding columns of the Llama 3.1 \verb.8b. and \verb.70b. embedding matrices.
We then take the cosine similarity of each pair.
We find increasing model size slightly increases pair similarity.
The average cosine similarity of pairs is 0.369 for \verb.8b. and 0.386 for \verb.70b., with the distance of the two distributions being $d\!=\!0.13$.
To verify that this effect is not the result of \emph{all} token embeddings becoming more similar as we increase model size, we calculate cosine similarity between 20,000 random pairs of token embeddings for \verb.8b. and \verb.70b..
In fact, the average pair of vectors is \textit{less} similar in the larger model: pairs sourced from \verb.8b. have a mean similarity of 0.016 versus 0.008 for \verb.70b. (Cohen's $d\!=\!-0.38$).

A t-SNE projection \citep{vandermaatenVisualizingDataUsing2008} of all LW/non-LW pairs in the embedding matrices of both Llama models is shown in \autoref{fig:tsne}.
Both projections integrate all tokens not belonging to a pair as a baseline.
Increasing model scale (and corresponding training time assuming compute optimality) results in vectors that t-SNE is better able to distribute uniformly.
We furthermore observe LW/non-LW tokens co-occupy space to a greater degree as scale increases, which suggests our measured cosine similarities suggest more exchangeable representations as model scale increases.
This indicates LW/non-LW token pairs are not wholly exchangeable on an embedding basis.

\paragraph{Whitespace in training data.}
Does the ratio of LW to non-LW tokens in pre-training data account for their variations in embedding space?
The actual pre-training data for these models is not available to us, but we can approximate its distribution using the Dolma dataset \citep{soldainiDolmaOpenCorpus2024}.
We use a sample of Dolma \verb,v1.6, containing 16GB of text taken from a variety of online sources.
We then tokenize Dolma once for each of the respective tokenizers for the four model families.
We first note our tokenizers differ in compression factor: the Llama, Gemma, and Qwen tokenizers converting Dolma into $\approx7.5*10^9$ tokens, while the smaller Phi vocabulary produces $\approx9.1*10^9$ tokens.

Counting token occurrences furthermore reveals LW tokens are consistently more prevalent than their non-LW token duplicates with all tokenizers.
We find the Llama, Gemma, and Qwen tokenizers emit LW tokens at rates from 2.5 to 2.7 times greater than non-LW tokens, indicating our models of interest were exposed to many more LW tokens than non-LW tokens during pre-training.
Phi is again an outlier, with its tokenizer producing LW tokens at a rate 1.66 times greater than non-LW tokens.
Our results provide one possible explanation for why incorporating LW into grammars leads to increased classification performance: the LW tokens were observed at a greater frequency during pre-training.
The Phi tokenizer producing fewer LW tokens relative to non-LW tokens supports this conclusion given we also observe Phi showing limited response to leading whitespace.

\section{Discussion}
Conditioning LLM results on specific output formats is a promising and popular way to ensure valid annotation results, but can reduce performance.
We systematically test a range of familiar formats that appear semantically similar to humans, along with a set of format options that users may not even be aware of.
We find there are substantial differences between seemingly identical formats.
By varying over tokens differentiated by syntactic (but not semantic) qualities, we show GCD-guided LLMs are sensitive to features like leading whitespace, numerical representation, and newlines.
By better evaluating and surfacing these differences, we hope to close the gap in output-filtered performance without modifying existing decoding systems.

We recommend the following strategies:

\paragraph{Evaluate multiple formats.}
We find that there is substantial variation between output formats and that while there are some consistent results, the effect of formats may vary between model families. 
As new models are released, developers and user communities should provide evidence-based guidance on models' preferred formats.

\paragraph{Use leading whitespace.}
Users may not notice that, when generating without structured output, models usually add leading whitespace.
Unintentionally forcing models to \textit{not} use an initial newline or tokens with leading spaces may prevent models from accessing well-trained token ranges.
The difference between an initial newline and space is less important, so long as whitespace is present.

\paragraph{Prefer the conventional format.}
We recommend selecting token ranges matching the intended output format.
If the task at hand requires numerical values be selected, then it is important the corresponding grammar incorporate numerical tokens.
Subword token representations in existing LLMs are not yet coherent enough to dependably exchange numerical tokens for their semantically-equivalent word tokens, or vice versa.
It is recommended to select context-appropriate patterns until models reach representation parity.

\paragraph{Use less-restrictive patterns.}
It is helpful to offer the model the greatest degree of choice if possible.
Consider a grammar guiding models to emit a scalar value --- our results suggest it is beneficial to offer the model a choice from the range $[0,100]$ over $[1,5]$.
It is further helpful to represent this range with the fewest tokens possible to prevent errors propagating over multiple inferences.

\paragraph{Use larger models}
Finally, our results indicate larger models, when available, are usually better.
They are generally (although not always) more capable and are typically less influenced by differences between output formats.

\section{Conclusion}

Models have token and formatting preferences that are not predictable nor transparent to users.
As trends move towards smaller local models --- and increasing GCD-based constraints to compensate --- this issue will become more pressing.
Going forward, these format preferences should be identified and reported on a model-by-model basis.

Grammar-constrained decoding (GCD) has the potential to close the gap between the task-specificity of fine-tuned models and the rapid iteration of zero-shot models. 
But users must not only write good prompts, they must also write effective grammars.
This work indicates that users may unintentionally knock models out of their ``comfort zone'' by inadvertently selecting patterns that are misaligned with pre-training or RLHF fine-tuning. We find that  understanding the effect of subtle format differences has significant benefits and opens new opportunities for research.

Advances in quantization, software infrastructure, and model training have made powerful LLMs increasingly available on laptop-grade hardware.
Of the eight LLMs studied in this work, six will load on computer systems containing 32GB of RAM.
The remaining two, Llama \verb.70b. and Qwen \verb.72b., require a minimum 48GB assuming quantization is kept to a minimum of six bits per weight.
We find smaller models are more likely to produce non-valid output, meaning GCD is especially valuable for researchers working in memory-constrained environments.
We also find that smaller models show more sensitivity to GCD formats.
The results of this work can help to ensure that the benefits of low-compute LLMs are fully realized.

The increasing popularity of structured output indicates there is appetite for employing LLMs as classifiers.
GCD is critical for ensuring the success of structured output, and we see our current brute-force method as a promising initial step in enabling consistent success.
Generating $1.28$ million generations is a time-consuming process.
Future work should consider more direct methods of assessing GCD optimality.
One pathway will be to study how training impacts subword token representations coherence.
Could modifying pre-training help ensure more consistent representations for similar tokens?

\section*{Limitations}
This work is limited in three respects.
First, the task patterns considered in our study target single scalar values, meaning more complex annotation tasks are not considered.
Second, we limit our study to so-called ``open-weight'' models whose weights are made freely available on the Internet with permissive licenses.
While our study compares models of varying sizes, including sizes matching those of proprietary models such as ChatGPT and Google Gemini, we leave open the question of whether proprietary models are similarly impacted by format variations such as leading whitespace.
Third, we do not provide conclusive evidence for a causal relationship between particular subword token representations and downstream classification performance.
We instead provide evidence with the goal of motivating future research on the relationship between token embeddings and LLM behavior.

\section*{Acknowledgements}
We would like to thank Axel Bax, Federica Bologna, Rebecca Hicke, Kiara Liu, Andrea Wang, Matthew Wilkens, and Shengqi Zhu for their kind comments and suggestions.

\bibliography{custom}

\appendix

\section{Example Task Prompts}
\label{sec:appendix}

We provide four example prompts, one from each of the benchmarks considered in this study.
All prompts are instantiations of the template described in \autoref{par:evaluation}.

\paragraph{STS Benchmark.}
This task has raters determine whether two texts are semantically similar.
Prompt:
\begin{verbatim}
<string 1>cooking</string 1>

<string 2>rice</string 2>

Rate your agreement with the following
statement: These strings are similar.

Respond only with a number between 0 and 1.
\end{verbatim}

Corresponding grammar:
\begin{verbatim}
root ::= response
response ::= "0."[0-9][0-9]
\end{verbatim}

\paragraph{MEN Test Collection.}
This task likewise has raters determine whether two texts are semantically similar.
Prompt:
\begin{verbatim}
<string 1>Obama praises U.S.-Latin America
trade ties</string 1>

<string 2>Obama Secret Service agents sent
home</string 2>

Rate your agreement with the following
    statement: These strings are similar.

Respond only with a number between 1 and 
5, where
1 = Strongly disagree,
2 = Disagree,
3 = Neither agree nor disagree,
4 = Agree,
5 = Strongly agree
\end{verbatim}

Corresponding grammar:
\begin{verbatim}
root ::= response
response ::= [1-5]
\end{verbatim}

\paragraph{Quora Duplicate Questions.}
This task has raters assess whether two strings are duplicates.
Prompt:
\begin{verbatim}
<string 1>Do ants die of old age?</string 1>

<string 2>Could plants die of old
age?</string 2> 

Rate your agreement with the following 
statement: These strings are duplicates. 

Respond only with 'True' or 
'False'.
\end{verbatim}

Corresponding grammar:
\begin{verbatim}
root ::= response
response ::= "True" | "False"
\end{verbatim}

\paragraph{ToxicChat.}
This task has raters assess whether there is toxic language present.
Prompt:
\begin{verbatim}
<string 1>Hi, I need to write a  snippet 
in javascript</string 1> 

<string 2>Sure, what do you need help 
with?</string 2>

Rate your agreement with the following 
statement: These strings contain toxic
language.

Respond only with a number 
between 1 and 5.
\end{verbatim}

Corresponding grammar:
\begin{verbatim}
root ::= response
response ::= [1-5]
\end{verbatim}

\end{document}